\documentclass{article}

\usepackage{arxiv}

\usepackage[utf8]{inputenc} 
\usepackage[T1]{fontenc}    
\usepackage{hyperref}       
\usepackage{url}            
\usepackage{booktabs}       
\usepackage{amsfonts}       
\usepackage{nicefrac}       
\usepackage{microtype}      
\usepackage{lipsum}		
\usepackage{graphicx}
\usepackage{natbib}
\usepackage{doi}

\usepackage{amsmath,amssymb,amsfonts}
\usepackage{algorithmic}
\usepackage{graphicx}
\usepackage{textcomp}
\usepackage{xcolor}
\usepackage{array}
\usepackage{booktabs}
\usepackage{bm}

\title{Attention-based Spatial-Temporal Graph~Neural~ODE for Traffic Prediction}

\author{ \href{https://orcid.org/0000-0002-7902-3568}{\includegraphics[scale=0.06]{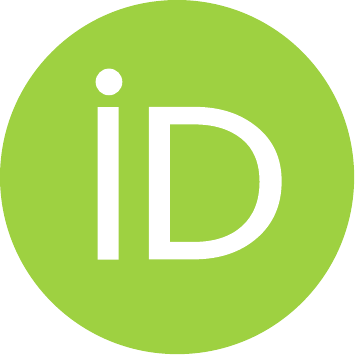}\hspace{1mm}Weiheng Zhong}\\
	Department of Civil and Environmental Engineering\\
	University of Illinois at Urbana-Champaign\\
	Champaign, USA \\
	\texttt{weiheng4@illinois.edu} \\
	\And
	\href{https://orcid.org/0000-0000-0000-0000}{\includegraphics[scale=0.06]{orcid.pdf}\hspace{1mm}Hadi Meidani} \\
	Department of Civil and Environmental Engineering\\
	University of Illinois at Urbana-Champaign\\
	Champaign, USA \\
	\texttt{meidani@illinois.edu} \\
        \And
	\href{https://orcid.org/0000-0000-0000-0000}{\includegraphics[scale=0.06]{orcid.pdf}\hspace{1mm}Jane Macfarlane} \\
	Sustainable Energy Systems Group\\
	Lawrence Berkeley National Laboratory\\
	Berkeley, USA \\
	\texttt{jfmacfarlane@lbl.gov} \\
}

\renewcommand{\shorttitle}{\textit{arXiv} Template}

\hypersetup{
pdftitle={A template for the arxiv style},
pdfsubject={q-bio.NC, q-bio.QM},
pdfauthor={David S.~Hippocampus, Elias D.~Striatum},
pdfkeywords={First keyword, Second keyword, More},
}

\begin{document}
\maketitle

\begin{abstract}
Traffic forecasting is an important issue in intelligent traffic systems (ITS). Graph neural networks (GNNs) are effective deep learning models to capture the complex spatio-temporal dependency of traffic data, achieving ideal prediction performance. In this paper, we propose attention-based graph neural ODE (ASTGODE) that explicitly learns the dynamics of the traffic system, which makes the prediction of our machine learning model more explainable. Our model aggregates traffic patterns of different periods and has satisfactory performance on two real-world traffic data sets. The results show that our model achieves the highest accuracy of the root mean square error metric among all the existing GNN models in our experiments. 
\end{abstract}


\section{Introduction}
\label{Section.Introduction}

Accurate traffic prediction is of significant importance for the intelligent transportation system, especially on the highway where a massive traffic flow of high driving speed is moving. Traffic forecasting is a long-standing challenge due to the complexity of spatio-temporal dependencies of traffic data. On the one hand, traffic time series have strong nonlinear temporal dynamics due to the varying traffic demand in different time points. On the other hand, the correlation between different locations in a network changes under different traffic conditions. Traffic accidents also cause non-stationary traffic dynamics, making the traffic dynamics more unpredictable. 

Graph neural networks (GNNs) attracted tremendous attention as a popular class of deep learning models for traffic prediction. Unlike fully-connected neural networks and convolution neural networks, GNNs can deal with graph-structured data. GNNs outperform most traditional methods by aggregating traffic information of neighbor locations and achieving satisfactory predictions of future traffic conditions. There are many excellent GNN models, such as DCRNN \cite{li2017diffusion}, STGCN \cite{yu2017spatio}, ASTGCN \cite{guo2019attention}, GraphWaveNet \cite{sun2018identifying}, STSGCN \cite{song2020spatial}, Graph Autoformer \cite{XAI}, Etc. These models have a specific framework to aggregate the spatial and temporal traffic features to predict future traffic data. 

Neural ODE \cite{chen2018neural} appears as an inspiration for us to predict the traffic future based on the dynamic of the traffic system. Spatial-temporal neural ODE \cite{zhou2021urban} combines neural ODE and convolution neural network for urban area traffic prediction, leveraging spatial-temporal data from multiple sources. Spatial-temporal graph neural ODE (STGODE) \cite{fang2021spatial} combines graph neural network and neural ODE to extract a more extended range of spatio-temporal correlation without being affected by the over-smoothing problem. However, to the best of our knowledge, there is not much literature that explicitly uses a deep learning model to learn the dynamics of the traffic system and predicts the future traffic data through the system's evolution.

In this paper, we focus on using neural ODE to imitate the dynamics of the traffic system by performing supervise learning on the output of neural ODE at each time step. We utilize attention mechanism \cite{feng2017effective} to model the dynamics of the traffic system, using the spatial and temporal attention to estimate the trend of traffic under different traffic conditions. We also shows the advantage of aggregating traffic trend of different periods for higher prediction accuracy. The adjoint training \cite{chen2018neural} effects are also discussed in this paper. 

\section{Problem Setup}
\label{Section.Problem_setup}

In this study, we define a traffic network as an undirected graph $G=(V,E,S)$, where $V$ is a set of all vertices and $E$ is a set of all edges of the network; $S \in \mathbb{R}^{N \times N}$ defines the adjacent matrix of the graph and $N=|V|$ is the number of all vertices. 

We use sensors in the vertices to collect measurements of the traffic features with a fixed sampling frequency within a period. The number of traffic features each sensor collects is $F$. Then the collected traffic data sequence $\bm{X} = \{ x_1, x_2,...,x_t,x_{t+1},...,x_T \}$, where $x_t \in \mathbb{R}^{N \times F}$ denotes the record of all the traffic features of all nodes at time $t$.

We denote $T_h$ as the time length of one hour, $T_d$ as the time length of one day, and $T_w$ as the time length of one week. Then we define four types of time series segments of the length of $T_h$:
\begin{itemize}
    \item The \textbf{predicted segment} $\bm{\chi}^{t_p}$ is defined as $\{ x_{t_p + 1}, x_{t_p + 2}, ..., x_{t_p + T_h} \}$, where $x_{t_p + 1}$ is the starting point of the predicted segment. We will use historical traffic data segments to predict this traffic segment.
    
    \item The \textbf{recent segment} $\bm{\chi}^{t_r}$ is defined as $\{ x_{t_p - T_h + 1}, x_{t_p - T_h + 2}, ..., x_{t_p} \}$, where $t_r = t_p - T_h$. It is the time segment that is temporally closest to the predicted segment, which provides us with most important information for the traffic condition of next one hour.
    
    \item The \textbf{daily period segment} $\bm{\chi}^{t_d}$ is defined as $\{ x_{t_p - T_d + 1}, x_{t_p - T_d + 2}, ..., x_{t_p - T_d + T_h} \}$, where $t_d = t_p - T_d$. This is the traffic data of the same time in a day as the predicted segment and we use it to capture the daily traffic pattern of the system such as the rush hour in the morning and evening. 
    
    \item The \textbf{weekly period segment} $\bm{\chi}^{t_w}$ is defined as $\{ x_{t_p - T_w + 1}, x_{t_p - T_w + 2}, ..., x_{t_p - T_w + T_h} \}$, where $t_w = t_p - T_w$. This is the traffic data of the same time in a week as the predicted segment and we use it to capture the weekly traffic pattern of the system. Using this time segment we aims at detecting the dynamic of the traffic system within the period of one week. 

\end{itemize}

\begin{figure}[ht]
    \label{fig.data_example}
    \centering
    \includegraphics[width=8cm]{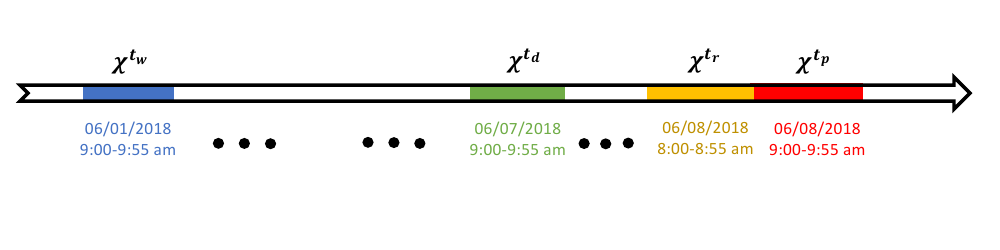}
    \centering
    \caption{\footnotesize An example of the input and output traffic segments of our model is shown. The recent traffic segment is defined as the traffic conditions of the previous hour. The daily traffic segment is the traffic conditions of yesterday at the same time slot as the predicted traffic segment. Similarly, The weekly traffic segment is the traffic conditions of last week at the same time slot of the same day as the predicted traffic segment.}
\end{figure}

We intend to build a deep learning model using the historical traffic data $\{ \bm{\chi}^{t_r}, \bm{\chi}^{t_d}, \bm{\chi}^{t_w} \}$ as inputs to predict future traffic data $\bm{\chi}^{t_p}$. An example of constructing input data and output data is shown in Figure \ref{fig.data_example}. 

\section{Methodology}
\label{Section.Methodology}

\subsection{Model framework}
\label{Subsection.Model_framework}

Our model consists of three independent Neural ODE blocks to capture the dynamic of the traffic system of different periods. We use a fully-connected network as our encoder to map the traffic feature data of each node in each time point to a higher-dimensional space, which is employed to strengthen the expressiveness of the model. 

Then we use Neural ODE block to calculate the hidden state of the future time segments by learning the traffic pattern of different time periods. We implement three times "Forward Pass" \cite{chen2018neural} of each Neural ODE block to obtain $\{H^{t+\Delta t}, H^{t + 2 \Delta t}, H^{t + 3 \Delta t}\}$, where $t$ is starting time point of the historical time segment and $\Delta t$ is the $\frac{1}{3}$ time length of the whole period. This means that $t_w + 3 \Delta t_w = t_d + 3 \Delta t_d = t_r + 3 \Delta t_r = t_p$. We consider $H^{t_{p,w}}, H^{t_{p,d}}, H^{t_{p,r}}$ as the hidden feature of predicted time segment of three independent ODE blocks.

\begin{figure}[ht]
    \label{fig.model_framework}
    \centering
    \includegraphics[width=6cm]{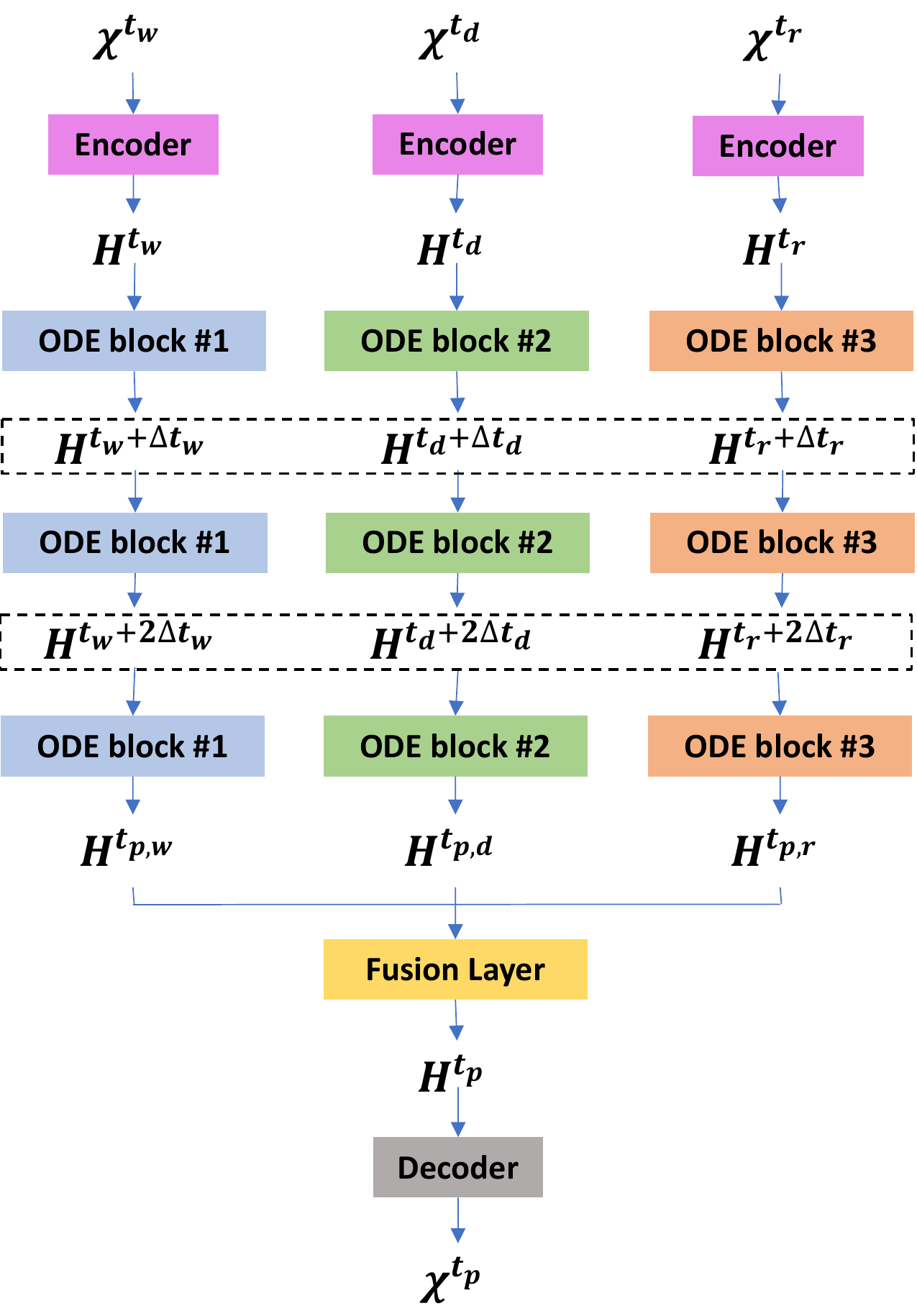}
    \centering
    \caption{\footnotesize The framework of the proposed attention-based spatial temporal graph Neural ODE is shown in this figure. We apply different ODE blocks to different input traffic segments, which are used to learn the traffic dynamics of different time intervals. A fully-connected fusion layer is applied to the aggregated features of three blocks and outputs the final aggregated features.}
\end{figure}

We use a fully connected neural network as a fusion layer to aggregate the hidden features $H^{t_{p,w}}, H^{t_{p,d}}, H^{t_{p,r}}$. The output of the fusion layer $H^{t_{p}}$ is the same dimension as each hidden state of $\{H^{t_{p,w}}, H^{t_{p,d}}, H^{t_{p,r}}\}$. As shown in Figure \ref{fig.model_framework}, we will output a set of intermediate hidden states $\mathbb{H} = \{\{H^{t_w+\Delta t_w}, H^{t_w+ 2 \Delta t_w}, H^{t_d+\Delta t_d}, H^{t_d+ 2 \Delta t_d}, H^{t_r+\Delta t_r}, $ $H^{t_r + 2 \Delta t_r}\}$ and hidden state of the predicted time segment $H^{t_p}$. We use a decoder, which is also a fully-connected neural network, as a mapping from the hidden state to the original data space. Combined with the decoder, We then can obtain the prediction of the predicted time segment $X^{t_p}$.

We perform supervised training for our model by minimizing the mean square error between predicted and ground-truth traffic data using stochastic gradient descent algorithms. The training loss $L$ is defined as:
\[
  L = MSE(\mathbb{X}, \mathbb{X}_{GT})
\]

where $X^{t_p}_{GT}$ are the ground-truth time segment data. The details of our model framework is shown in Figure \ref{fig.model_framework}.

\subsection{Neural ODE block}
\label{Subsection.NODE_modules}

Each ODE block in Figure \ref{fig.model_framework} consists of a spatial attention layer, a temporal attention layer, and a chebyshev graph convolution layer. 

The traffic conditions of one location can affect the traffic condition of other neighbor locations. For this consideration, we adopt the chebyshev graph convolution to aggregate the information of the neighbor vertices \cite{kipf2016semi}. In order to allow for capturing multi-hop neighbors' traffic conditions, we use up to third order of chebyshev polynomials. 

\begin{figure}[ht]
    \label{fig.ODE_block_framework}
    \centering
    \includegraphics[width=8cm]{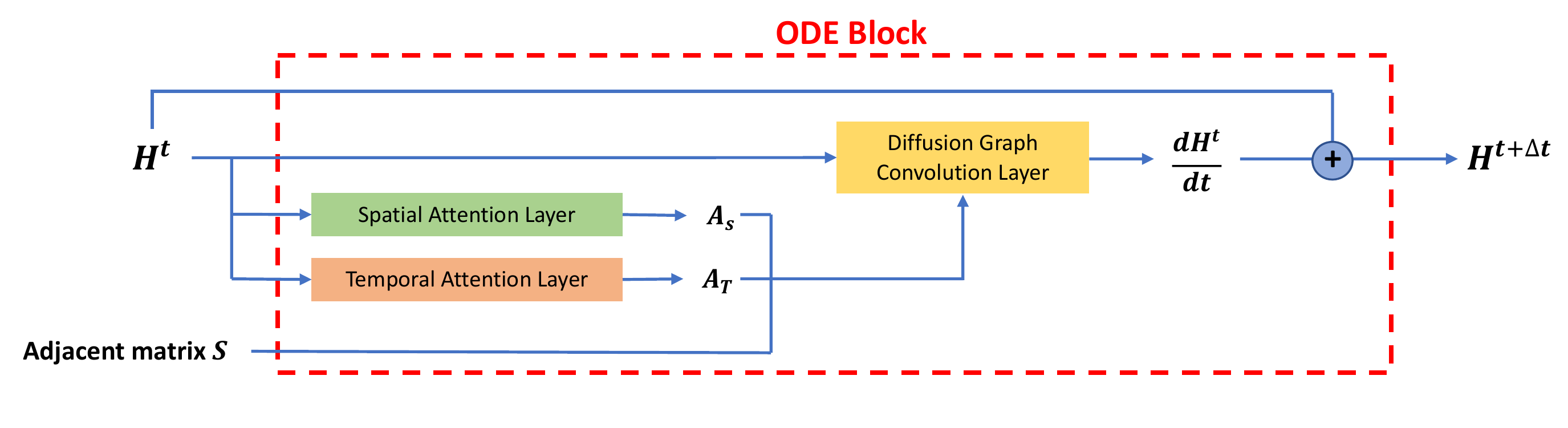}
    \centering
    \caption{\footnotesize The architecture of an individual neural ODE block is shown. We will compute the spatial attention scores and temporal attention scores based on the input traffic features. Combined with exact graph connectivity, we perform diffusion graph convolution to derive the traffic features of the next time step.}
\end{figure}

However, the influence of neighbor vertices is highly dynamic. We may need to pay different attention to the same neighbor vertices under different traffic conditions when we aggregate the information of the neighbor vertices. Also, in the temporal dimension, correlations between traffic conditions in different time slices vary in different situations. Hence, we use the attention mechanism \cite{feng2017effective} to capture spatial and temporal correlations by calculating the attention scores $A_S$ and $A_T$, using the spatial and temporal attention layers. We couple all information in chebyshev graph convolution layer to predict the change of traffic condition $\frac{d H^t}{dt}$ over a specific time length $\Delta t$. The output of the ODE block is the hidden state of future time segment $H^{t+\Delta t} = H^{t} + \frac{d H^t}{dt}$. The details of the ODE block architecture is shown in Figure \ref{fig.ODE_block_framework}.

\section{Numerical Experiments}
\label{Section.Experiments}

\subsection{Data set description}
\label{Subsection.Dataset}

We use two real-world traffic data sets, PeMS-BAY \cite{li2017diffusion} and PeMS04 \cite{guo2019attention}, to validate our model. These two data sets are collected by the Caltrans Performance Measurement System (PeMS) every 30 seconds [STGODE]. Both the traffic data are aggregated into 5-minutes intervals. In PeMS-BAY data set, we only have the data of traffic flow velocity. There are 325 sensors in Bay Area collecting six months of data from Jan $1^{st}$ 2017 to May $31^{st}$ 2017. As for PeMS04 data set, we have 307 sensors on the highway of major metropolitan areas in California measuring three traffic features, including traffic flow, average speed, and average occupancy, for almost two months. 

We use our model to predict traffic speed using PeMS-BAY data set and traffic flow using PeMS04 data set. Both data sets are sorted by time from the past to the present. We split the data into three parts for training (70\%), validation (10\%), and testing (20\%). Data normalization with the mean and standard deviation of the training data is applied to these three parts. 

\subsection{Baseline models}
\label{Subsection.Baselines}

We compare our model performance with several baseline models:
\begin{itemize}
    \item Historical average (HA) \cite{ermagun2018spatiotemporal}: it estimates seansonal traffic pattern and uses weighted average of traffic flow data as prediction. 
    
    \item Auto-regressive integrated moving average model (ARIMA) \cite{ermagun2018spatiotemporal}: it is a traditional parametric for analysis of time series data.
    
    \item Fully connected LSTM (FC-LSTM) \cite{sutskever2014sequence}: it is an RNN-based sequence model whose encoder and decoder are both LSTM layers.
    
    \item Diffusion convolutional recurrent neural network (DCRNN) \cite{li2017diffusion}: it purposes a combination of diffusion convolution operator and GRU \cite{fu2016using} to capture spatio-temporal correlations.
    
    \item Spatial-temporal graph ODE (STGODE) \cite{fang2021spatial}: aggregating the spatial and temporal information of the traffic segments, it uses Neural ODE to learn the system's dynamic.
    
    \item Attention based spatial temporal graph convolutional networks (ASTGCN) \cite{guo2019attention}: it uses attention mechanism to capture spatial and temporal data dependency for better future prediction.
    
    \item Graph multi-attention network (GMAN) \cite{zheng2020gman}: it uses a multi-head self-attention mechanism to capture the spatial and temporal correlation of traffic series data and transform attention to capture dependency between historical time segments and future time segments. 
\end{itemize}

\subsection{Comparison of traffic prediction performance}
\label{Subsection.Model_performance}

We use two different metrics to evaluate the model performance: the root mean squared error (RMSE) and mean absolute error (MAE). We will exclude missing data when evaluating the performance of both data sets. Using Adam optimizer, we use a learning rate of 0.0001 and train our model for 50 epochs. The tradeoff coefficient $\alpha$ is set to be 0.1, and the batch size is set to be 32.

We report the traffic prediction error of 15-min ahead prediction, 30-min ahead prediction, and 60-min ahead prediction in Table \ref{Table.speed_prediction_performance}, using PeMS-BAY data set for validation. We observe that our method outperforms all the baseline models in 15-min ahead prediction and 30-min ahead prediction. For 60-min ahead prediction, our model achieves significant improvement in decreasing RMSE of the prediction and slight improvement in decreasing MAE. These results demonstrate the effectiveness of learning traffic dynamics to predict future traffic conditions. 

\begin{table}[ht]
\label{Table.speed_prediction_performance}
\caption{Traffic speed prediction performance of different models on PeMS-BAY data set}
\centering
\begin{tabular}{c c c c c c c }
\hline
model & \multicolumn{2}{c}{15 min} & \multicolumn{2}{c}{30 min} & \multicolumn{2}{c}{60 min} \\

{} & RMSE & MAE & RMSE & MAE & RMSE & MAE \\
\hline
HA & 5.60 & 2.88 & 5.60 & 2.88 & 5.60 & 2.88 \\
ARIMA & 3.30 & 1.61 & 4.76 & 2.33 & 6.50 & 3.39\\
FC-LSTM & 4.19 & 2.05 & 4.55 & 2.2 & 4.96 & 2.36 \\
DCRNN & 2.95 & 1.38 & 3.97 & 1.74 & 4.74 & 2.05 \\
ASTGCN & 2.80 & 1.35 & 3.82 & 1.66 & 4.56 & 2.06 \\ 
GMAN & 2.82 & 1.34 & 3.72 & 1.62 & 4.32 & \textbf{1.86} \\
{Our method} & \textbf{2.68} & \textbf{1.30} & \textbf{3.36} & \textbf{1.62} & \textbf{4.10} & 2.00 \\
\hline
\end{tabular}
\end{table}

We also compare our model's performance on the average results of traffic flow prediction using PeMS04 data set over the next one hour in Table \ref{Table.flow_prediction_performance}. The results show that our model also outperforms other baseline models in traffic flow prediction. Neural ODE-related model (STGODE and our model) has higher performance than other baseline models. Moreover, it also shows that our model has good performance in decreasing RMSE of the prediction.

\begin{table}[ht]
\label{Table.flow_prediction_performance}
\caption{Traffic flow prediction performance of different models on PeMS04 data set}
\centering
\begin{tabular}{c c c}
\hline
model & RMSE & MAE \\
\hline
HA & 54.11 & 36.76 \\
ARIMA & 68.13 & 32.15\\
FC-LSTM & 45.76 & 29.50 \\
DCRNN & 37.61 & 24.63 \\
ASTGCN & 35.22 & 22.93 \\ 
STGODE & 32.82 & 20.84 \\
{Our method} & \textbf{30.65} & \textbf{20.23} \\
\hline
\end{tabular}
\end{table}

\subsection{Effect of the fusion layer}
\label{Subsection.effect_fusion_layer}

We consider every component in our model has some specific physical meaning except for the fusion layer. The encoder and decoder are two opposite mappings between observed traffic data and the hidden state of the dynamic traffic system. Neural ODE imitates the evolution of the dynamic system. However, it is hard to explain the output of the fusion layer. So we will investigate the effect of the fusion layer in our model.

\begin{figure}[ht]
    \label{fig.period_effect}
    \centering
    \includegraphics[width=8cm]{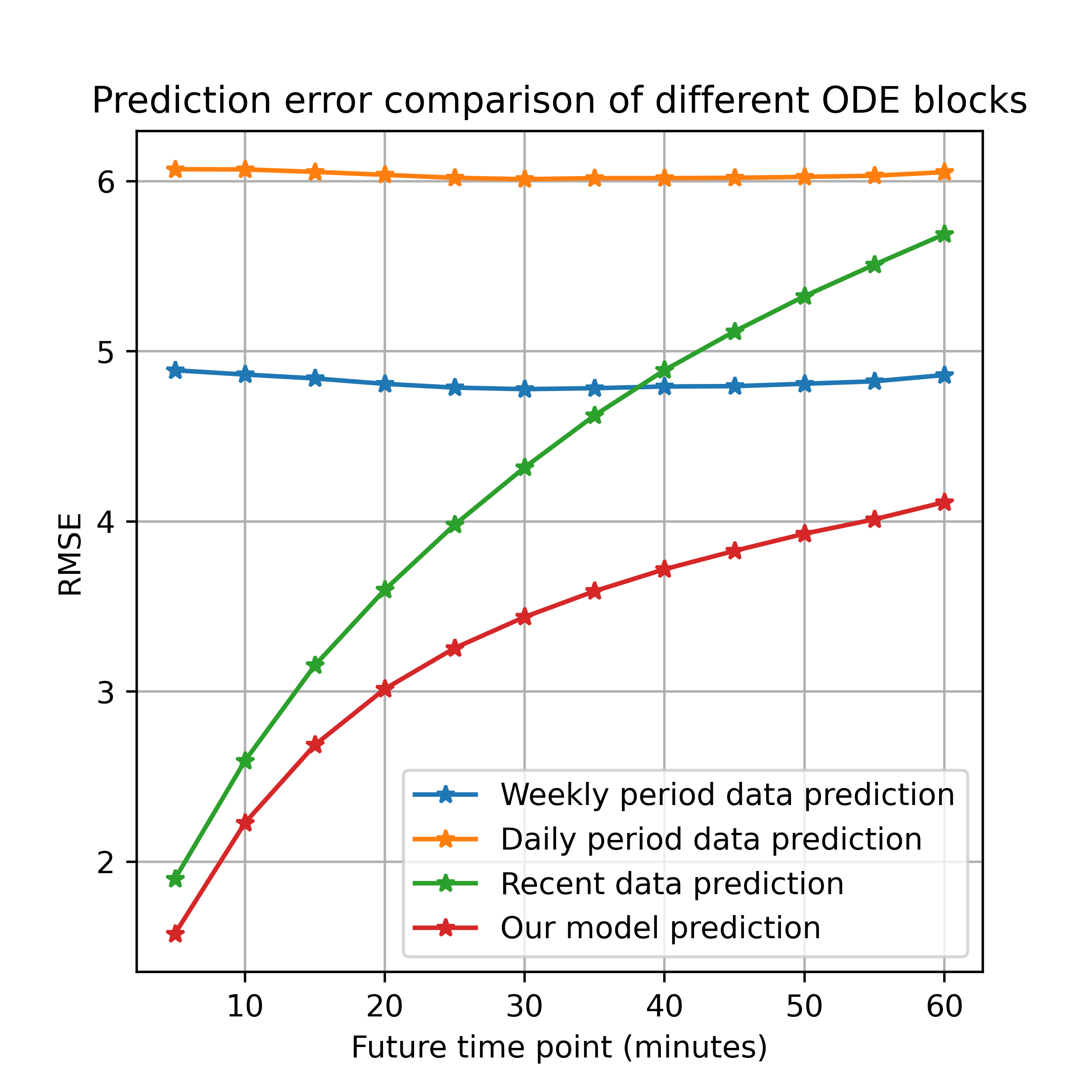}
    \centering
    \caption{\footnotesize Prediction error comparison of different neural ODE blocks and our purposed model. We used RMSE to compare the performance of using different traffic features. We observed that weekly period segment was useful for long-term future prediction and recent traffic segment was beneficial for short-term future prediction. The performance of our model showed the effectiveness of the fusion layer to achieve higher performance in our model.}
\end{figure}

We remove the fusion layer and use the prediction of three independent ODE blocks for comparison. We consider that dynamic of different periods will have different effects on our prediction. As shown in Figure \ref{fig.period_effect}, recent traffic data has a significant effect on short-term traffic prediction (5min-20min), while weekly period data is more important in long-term traffic prediction (45min-60min). The ODE block of daily period data does not have ideal short-term or long-term prediction performance. We consider that daily traffic pattern is more difficult to capture due to abrupt change in rush hour traffic conditions. However, by coupling features of different traffic patterns, we can leverage different historical data effects on the prediction of different time points, which helps us increase the model accuracy of future traffic prediction.

\subsection{Effect of adjoint training method}
\label{Subsection.effect_adjoint_training}

The adjoint training method highly reduces the memory cost of Neural ODE training \cite{chen2018neural}. However, Some researchers claimed that the adjoint training method of Neural ODE introduces error into gradient information, which will negatively influence the final well-trained model \cite{daulbaev2020interpolated}. Here we also implement two training methods, adjoint training and regular training. Then we compare the performance of models trained by these two methods.

In Table \ref{Table.effect_of_adjoint}, we conclude that the adjoint training method decreases prediction accuracy, and GMAN has higher accuracy in MAE if we use the adjoint method to train our model. However, the model of adjoint training still outperforms GMAN in the metric of RMSE.

\begin{table}[ht]
\label{Table.effect_of_adjoint}
\caption{Effect of adjoint training on prediction error.}
\centering
\begin{tabular}{c c c c c c c}
\hline
model & \multicolumn{2}{c}{15 min} & \multicolumn{2}{c}{30 min} & \multicolumn{2}{c}{60 min} \\

{} & RMSE & MAE & RMSE & MAE & RMSE & MAE \\
\hline
GMAN & 2.82 & 1.34 & 3.72 & 1.62 & 4.32 & \textbf{1.86} \\
{adjoint training} & 2.72 & 1.33 & 3.52	& 1.72 & 4.22 & 2.05\\
{non-adjoint training} & \textbf{2.68} & \textbf{1.30} & \textbf{3.36} & \textbf{1.62} & \textbf{4.10} & 2.00 \\
\hline
\end{tabular}
\end{table}

\begin{figure}[ht]
    \label{fig.adjoint_effect}
    \centering
    \includegraphics[width=8cm]{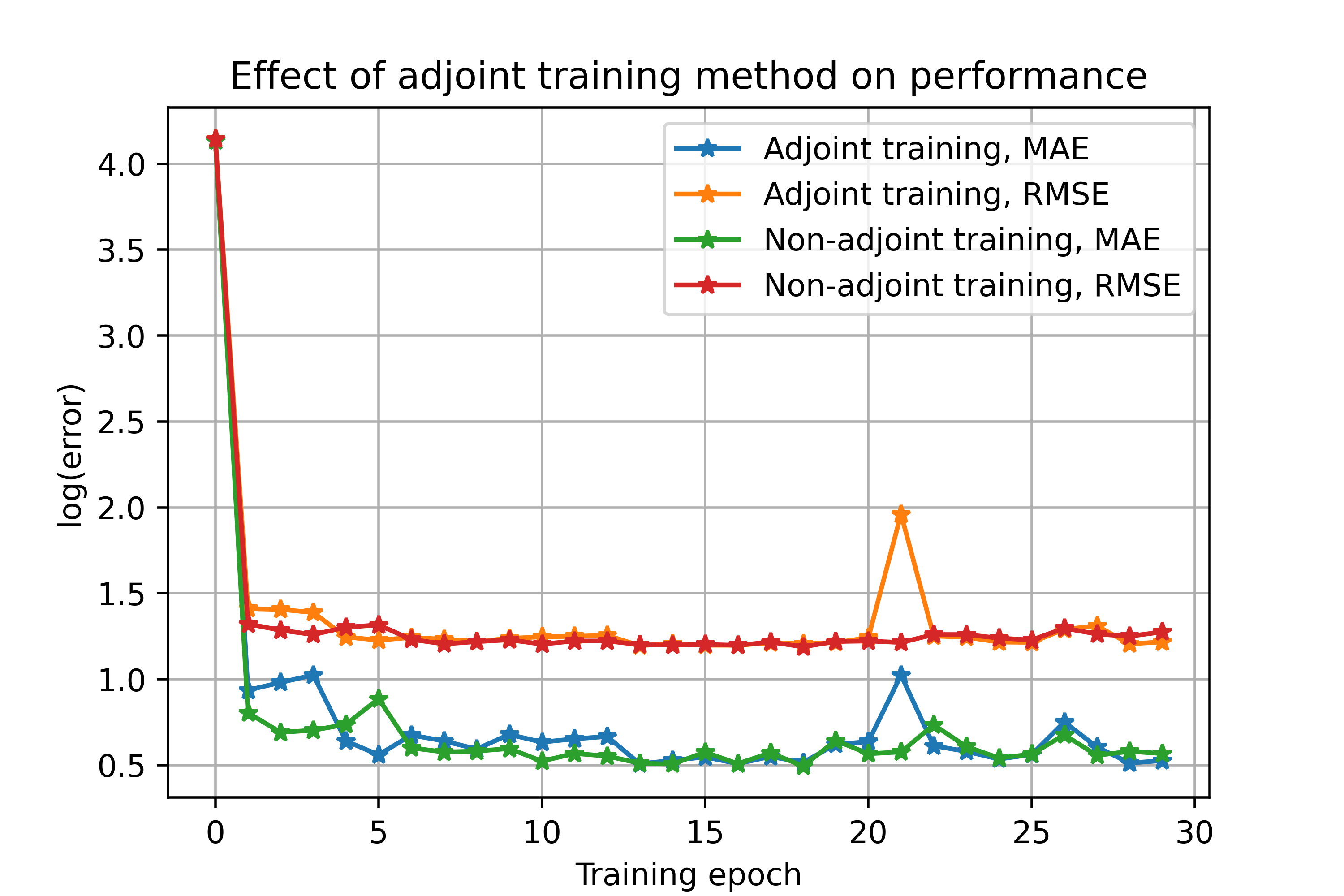}
    \centering
    \caption{\footnotesize The effect of the adjoint method on training stability is shown. We compared the error evolution over the training process under different training methods. We observed that using adjoint training slightly reduced the stability of the model training. However, it will not affect the final model performance if we train the model with enough epochs.}
\end{figure}

We also plot the average prediction error of our models in each training epoch in Figure \ref{fig.adjoint_effect}. We can observe that the prediction error of the adjoint training model has more significant vibration, which indicates that adjoint training is less stable than non-adjoint training. 

Considering the factors of model performance and training stability, it is suggestive that not use adjoint training unless dealing with an extremely large neural ODE model. 

\section{Conclusion}
\label{Section.conclusion}

In this paper, we propose a novel deep learning architecture to capture the dynamics of the traffic network system. We help our Neural ODE better imitate the evolution of the traffic system by introducing the attention mechanism to capture the spatial and temporal correlation between traffic data. We also purpose a fusion layer to aggregate features of different periodic dynamics to perform more accurate predictions for the future traffic data. The results show that our model can outperform most of the existing models in the root mean square error metric. 

However, the attention mechanism is not the exact dynamics of the traffic system. To make the model more explainable, we can introduce more physics-related components in the neural ODE block, such as diffusion processes. Also, adjoint training is an efficient computation method of training our model. If we can increase the stability of the adjoint training, we can construct a more complicated deep learning model to capture the temporal dependency better.

\bibliographystyle{unsrtnat}
\bibliography{references}

\end{document}